\documentclass[12pt]{article}

\usepackage{authblk}
\usepackage{lmodern}
\usepackage[T1]{fontenc}
\usepackage[utf8]{inputenc}
\usepackage[margin=1in]{geometry}
\usepackage{amssymb,amsmath,amsthm,amsfonts}

\usepackage{xcolor}
\usepackage{soul} 
\usepackage{tikz}
\usepackage{pgfplots}
\usepackage{makecell}
\usepackage[inline]{enumitem}
\usepackage{subcaption}
\usepackage{graphicx}
\usepackage{multirow}

\usepackage{pifont}

\usetikzlibrary{arrows,shapes,positioning,shadows,trees}

\tikzset{
  basic/.style  = {draw, text width=2cm, drop shadow, font=\sffamily, rectangle},
  root/.style   = {basic, rounded corners=2pt, thin, align=center,
                   fill=green!30},
  level 2/.style = {basic, rounded corners=6pt, thin,align=center, fill=green!60,
                   text width=8em},
  level 3/.style = {basic, thin, align=left, fill=pink!60, text width=6.5em}
}

\usepackage[pagebackref]{hyperref}
\hypersetup{
    colorlinks,
    linkcolor={red!50!black},
    citecolor={blue!50!black},
    urlcolor={blue!80!black},
    pdfstartview={XYZ null null 1.25}
}

\providecommand{\keywords}[1]
{
  \textbf{\textit{Keywords---}} #1
}

\def\mySectionSymbol~{\S{}}

\begin{document}

\title{Deep Learning-Based Approaches for Contactless Fingerprints Segmentation and Extraction}

\newcommand{\clrksn}{{$^\mathsection$}}
\newcommand{\msft}{{$^\dagger$}}
\newcommand{\srcinc}{{$^{\ddagger\ddagger}$}}
\newcommand{\wisconsin}{{$^\dag$}}

\renewcommand*{\Authsep}{\ }
\renewcommand*{\Authand}{\ }
\renewcommand*{\Authands}{\ }

\author[\wisconsin]{M.G. Sarwar Murshed}
\author[\clrksn]{Syed Konain Abbas} 
\author[\clrksn]{Sandip Purnapatra}
\author[\clrksn]{\newline Daqing Hou} 
\author[\clrksn]{Faraz Hussain} 
\affil[\wisconsin]{University of Wisconsin–Green Bay, WI, USA\authorcr {\tt \{murshedm\}@uwgb.edu}\vspace{0.4em}}
\affil[\clrksn]{Clarkson University, Potsdam, NY, USA\authorcr {\tt \{abbas,purnaps,dhou,fhussain\}@clarkson.edu}\vspace{0.4em}}


\date{}
\maketitle

\begin{abstract}
Fingerprints are widely recognized as one of the most unique and reliable characteristics of human identity. Most modern fingerprint authentication systems rely on contact-based fingerprints, which require the use of fingerprint scanners or fingerprint sensors for capturing fingerprints during the authentication process. Various types of fingerprint sensors, such as optical, capacitive, and ultrasonic sensors, employ distinct techniques to gather and analyze fingerprint data. This dependency on specific hardware or sensors creates a barrier or challenge for the broader adoption of fingerprint based biometric systems. This limitation hinders the widespread adoption of fingerprint authentication in various applications and scenarios. Border control, healthcare systems, educational institutions, financial transactions, and airport security face challenges when fingerprint sensors are not universally available.\par 
To mitigate the dependence on additional hardware, the use of contactless fingerprints has emerged as an alternative. Developing precise fingerprint segmentation methods, accurate fingerprint extraction tools, and reliable fingerprint matchers are crucial for the successful implementation of a robust contactless fingerprint authentication system. This paper focuses on the development of a deep learning-based segmentation tool for contactless fingerprint localization and segmentation. Our system leverages deep learning techniques to achieve high segmentation accuracy and reliable extraction of fingerprints from contactless fingerprint images. In our evaluation, our segmentation method demonstrated an average mean absolute error (MAE) of 30 pixels, an error in angle prediction (EAP) of 5.92 degrees, and a labeling accuracy of 97.46\%.
These results demonstrate the effectiveness of our novel contactless fingerprint segmentation and extraction tools. 
\end{abstract}

\keywords{Contactless, Deep learning, Segmentation, Fingerprints, Biometrics }

\section{Introduction}\label{sec:chp_8_introduction}
Biometric recognition systems play important roles in different domains due to their accuracy and quick processing time. Biometric authentication is based on an individual's distinct physiological or behavioral traits, such as voice prints, iris patterns, or fingerprints \cite{oostdijk2016state}. It reduces the risk of unauthorized access and burdens like the requirement of memorizing passwords, fraudulent authentication, and password stealing, as most of the biometric traits are difficult to forge or replicate, providing a higher level of security compared to passwords, which can be easily guessed or stolen unless the password is strong. Therefore, such authentication systems are becoming increasingly common in situations where users need to quickly authenticate themselves, such as border crossings, airports, cell-phone based authentication, and in healthcare, by simply providing their biometric data \cite{neal2016surveying}. This eliminates the need for complex login procedures or repetitive authentication steps.

One of the most popular and extensively utilized biometric characteristics for authentication is fingerprints. Each fingerprint is unique and remains consistent over time. To enhance security and accuracy, multiple fingerprints, such as the fingerprint slap, are used for a user instead of a single fingerprint. The first step in using a slap fingerprint to authenticate a user is to separate or segment each finger in the slap \cite{cadd2015fingerprint}. There are several publicly and commercially available slap fingerprints segmenters, such as NIST NFSEG \cite{ko2007users} and Neurotechnology Verifinger Segmenter \cite{grosz2023afr}.\par
We have developed a deep learning-based fingerprint segmenter called CRFSEG \cite{murshed2023deep} which outperforms other fingerprint segmenters in terms of precise fingerprint detection and fingerprint matching accuracy. However, it is important to note that these fingerprint segmenters have mainly been developed and used for contact-based slap fingerprint images.
While contact-based slap fingerprint authentication offers significant advantages, it also has multiple limitations, including the requirement for specialized hardware or infrastructure for implementation.

Most fingerprint authentication systems require physical contact between the fingers and fingerprint-capturing devices to capture fingerprint images. However, such contact can result in problems like blurred images due to dust on the capturing surface of the device and the potential for contamination from previously captured fingerprint information. Furthermore, fingerprint data can raise privacy concerns, and there is a possibility of false positives or false negatives during the matching process.
In addition to these limitations, contact-based fingerprint-capturing processes raise health concerns, particularly during pandemics, as they require direct contact with capturing devices \cite{priesnitz2022mobile}.
\begin{figure}
 \centering
 \includegraphics[width=0.4\linewidth]{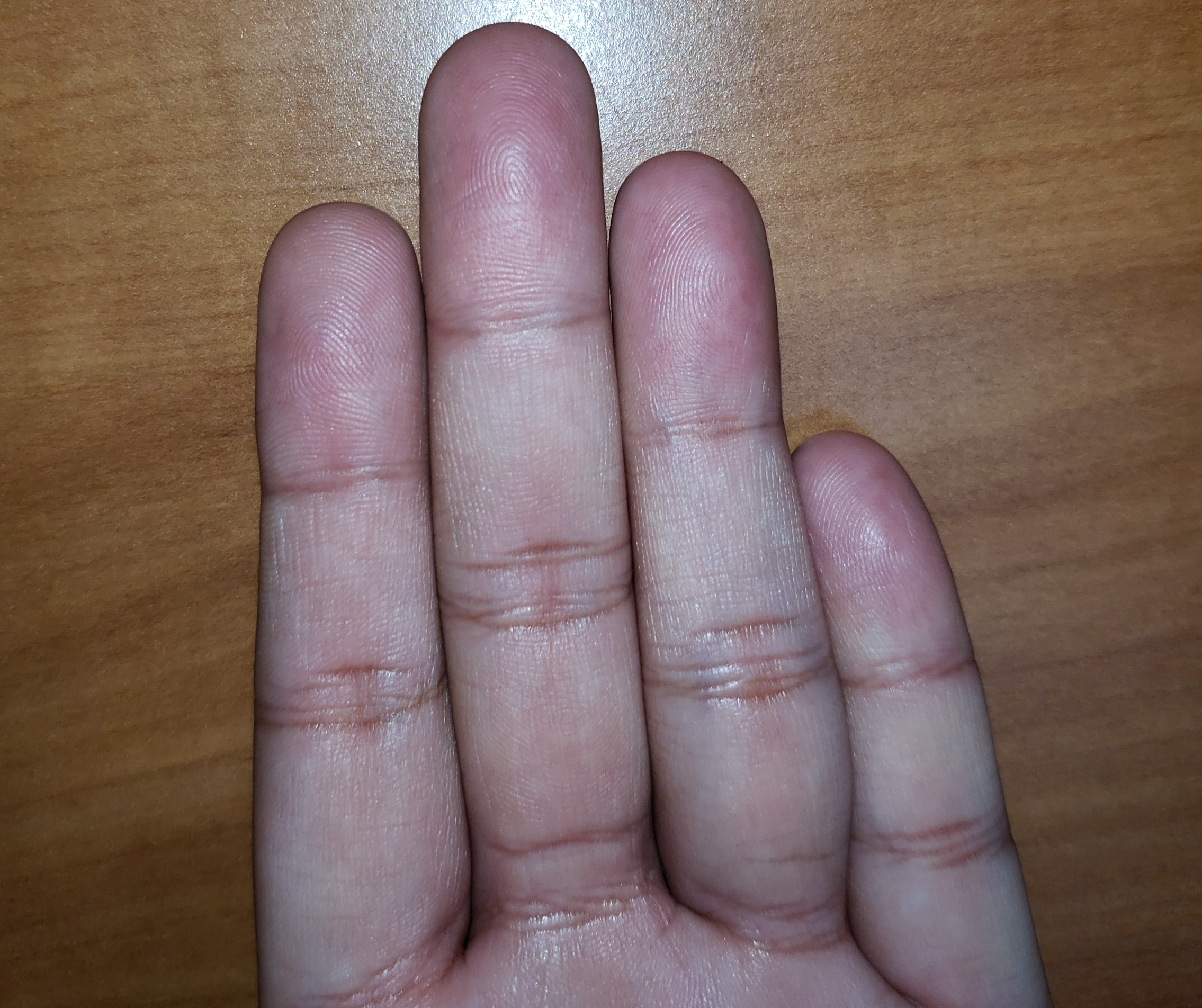} 
 \caption{A sample fingerphoto image was taken with a simple mobile phone camera. Fingerphotos are typically acquired by capturing an image of human fingers using a standard smartphone camera, and they frequently include multiple fingers within the frame.}
 \label{fig:fingerphoto}
\end{figure}
However, ongoing advancements in fingerprint-based biometric technology continue to address these challenges, making biometrics a promising and increasingly adopted authentication solution. One promising area is the use of contactless fingerprinting, which eliminates the need for additional specialized hardware or infrastructure. The term \textit{fingerphoto} refers to a contactless fingerprint image that may be taken with any camera, including a simple mobile phone camera. A fingerphoto is created by capturing an image of human fingers with a basic smartphone camera and typically includes multiple fingers \cite{9672054}. In \autoref{fig:fingerphoto}, a sample fingerphoto is shown.

Contactless fingerprint-based authentication systems offer several advantages over traditional contact-based fingerprint authentication systems:

\begin{enumerate}[label=\itshape\roman*\upshape)]

\item Improved security against spoofing: One of the most important advantages of using contactless fingerprint biometric systems is that they often incorporate additional anti-spoofing measures to enhance security. Advanced sensors and algorithms can detect and differentiate between live fingers and spoofing attempts using fake or synthetic fingerprints, reducing the risk of unauthorized access. 
\item Hygiene and cleanliness:  By eliminating the requirement of physical contact between the user's finger and the fingerprint-capturing sensor, contactless fingerprint authentication reduces the risk of spreading germs, viruses, and bacteria, making it a more hygienic option, especially in high-traffic areas or shared devices.

\item Convenience: Contactless fingerprint authentication is often faster and more user-friendly than contact-based methods. Users can simply place their finger toward the sensors/cameras or the camera near the user's hand, eliminating the need for precise alignment or direct physical contact. Most contactless fingerprint sensors like Idemia MorphoWave, Fujitsu PalmSecure can be used from a distance, which can be helpful for people with disabilities or who have difficulty bending over. This improves the overall user experience and can lead to increased user adoption and satisfaction.

\item Versatility and ease of integration: Contactless fingerprint biometric systems can be integrated into a wide range of devices and environments. They can work with existing touchless technologies, such as proximity sensors or facial recognition, to provide multi-modal authentication options. This versatility allows for seamless integration into various applications, including access control systems, smartphones, and payment terminals.



\end{enumerate}

Despite the advantages offered by contactless fingerprint biometrics, it is worth noting that they also have some limitations. These include potential susceptibility to environmental factors such as lighting conditions, distance captures, and uneven focus, as well as the need for appropriate sensor technology. The implementation of multispectral imaging technology exemplifies a suitable approach to address certain challenges associated with contactless fingerprint biometrics. Multispectral fingerprint sensors excel in overcoming environmental obstacles by capturing data across multiple wavelengths of light enabling them to mitigate issues arising from factors such as lighting conditions and other environmental variables. Advancements in sensor technology and algorithmic improvements continue to help overcome these challenges and make contactless fingerprint biometrics a promising and viable authentication solution \cite{bartolic2022fluorescence}.

Most contactless fingerprint authentication systems utilize fingerphotos, which are contactless fingerprint images that are captured from multiple fingers. Segmenting all fingertips from fingerphotos is an active area of research in contactless biometric authentication systems. Segmentation plays a crucial role in fingerprint matching, as observed in the existing literature \cite{maltoni2022fingerprint}.\par 

In this paper, we describe our novel contactless fingerprint segmentation system developed by enhancing our existing contact-based fingerprint segmentation model, CRFSEG (Clarkson Rotated Fingerprint Segmentation Model) \cite{murshed2023deep}, by updating its architecture and training it on a contactless dataset. Our novel contactless segmentation model CRFSEG-v2 demonstrates higher accuracy when evaluated on our in-house contactless fingerprint dataset consisting of 23,650 slaps, which were annotated by human experts. We describe a novel model for segmentation of contactless fingerphotos. Its architecture is based on our prior fingerprint segmentation system \cite{murshed2023deep}. Special attention was given to optimizing the deep learning architecture for this purpose. The paper presents the following novel contributions:

\begin{itemize}
    \item Built an in-house contactless dataset that contains 23,650 fingerphotos (94,600 single fingers).
    \item Annotated all fingerphoto images\footnote{The total number of images is 23,650. Out of 23,650, 2150 fingerphotos were human annotated and 21,500 were generated by augmentation method.} manually to establish a ground-truth baseline for the accuracy assessment of fingerprint segmentation systems.
    \item Updated and retrained for contactless fingerphotos our previously developed age-invariant deep learning-based slap segmentation model, that can handle arbitrarily orientated fingerprints.
    \item Assessed the performance of the contactless model named CRFSEG-v2 (Clarkson Rotated Fingerprint Segmentation Model) using the following metrics, MAE, EAP, and accuracy in fingerprint matching. 
\end{itemize} 

\section{Related Work}\label{sec:chp_8_related_work}
Researchers have utilized various segmentation methods to achieve precise contactless fingerprint segmentation. These methods employ different algorithms, including convolutional neural networks (CNN), recurrent adversarial learning, fuzzy C-mean (FCM), and genetic algorithm (GA) \cite{grosz2021c2cl}.
Wild et al. introduced a Skin-Mask finger segmentation technique to segment contactless finger photos \cite{Wild2019ComparativeTO}. The study employed a filtering approach utilizing the COTS fingerprint matcher (Verifinger) and the NFIQ 2.0 quality metric. They used a dataset comprising 2582 contact-based and 1728 contactless images from 108 fingers, resulting in TAR values ranging from 95.5\% to 98.6\% at FAR=0.1\%. It involves reducing the image resolution, creating masks based on the HLS (Hue, Lightness, and Saturation) color space, and selecting finger pixels based on correlation values. The technique also includes morphological operations, variance calculations, and contour extraction to obtain individual finger images.
However, the skin-mask finger segmentation technique relies on the assumption that the skin color of the fingers is relatively uniform. This assumption may not hold true in all cases, such as when the fingers are dirty or have been exposed to sunlight. This type of segmentation is sensitive to environmental factors such as lighting conditions and variations in skin color. The presence of objects or background elements that resemble skin tones may affect the accuracy of the segmentation results. Additionally, the technique may not include all finger pixels in the selection, and the separation of fingers connected via the thumb relies on evaluating local minima and maxima, which may not always be accurate.

Malhotra et al. introduced a method for segmenting the distal phalange in contactless fingerprint images by combining a saliency map and a skin-color map \cite{9107238}. They adopted a different approach by employing a random decision forest matcher and feature extraction with a deep scattering network. Their dataset included 1216 contact-based and 8512 contactless images from 152 fingers, resulting in an EER ranging from 2.11\% to 5.23\%. This process involves extracting a binary mask that represents the finger region in a captured finger selfie. It combines region covariance-based saliency and skin color measurements for effective segmentation. The steps include extracting visual features, constructing covariance matrices, computing dissimilarities between regions, generating saliency maps, converting to the CMYK (Cyan(C), Magenta(M), Yellow(Y), and BlacK(B)) color model, fusing saliency and skin color maps, and applying thresholds to obtain the final segmented mask. Although the algorithm produces impressive results, it requires extensive tuning of hyperparameters and it continues to struggle with accurately distinguishing fingerprints in the presence of noisy backgrounds or under excessively bright lighting conditions.

To address challenges in accurately segmenting fingerprints under challenging illumination conditions or noisy backgrounds, Grosz et al. an auto-encoder-based segmentation approach \cite{grosz2021c2cl}. To perform 500 PPI deformation and scale correction on contactless fingerprints, they utilized a spatial transformer. Their dataset included three parts one with 8,512 contactless and 1,216 contact-based fingerprints from 152 fingers, another containing 2,000 contactless and 4,000 contact-based fingerprints from 1,000 fingers, and a ZJU dataset comprising 9,888 contactless and 9,888 contact-based fingerprints from 824 fingers. These datasets served as the basis for evaluating their methodology. Their approach achieved impressive EERs of 1.20\%, 0.72\%, 0.30\%, and 0.62\% on these datasets, respectively. A U-net segmentation network is employed to segment a distal phalange of contactless fingerprint photos. The segmentation network is trained using manually marked segmentation masks from a dataset. The segmentation algorithm takes unsegmented images as input and outputs a segmentation mask. This mask is then used to crop a distal phalange of fingerprints and the background is removed. Image enhancements, including histogram equalization and gray-level inversion, are applied to improve ridge-valley structure. 

None of the aforementioned research studies provide information about the accuracy like mean absolute error (MAE), error in angle prediction (EAP), and labeling accuracy of their finger photo segmentation techniques. Our work addresses this gap by evaluating our novel contactless segmentation system using a large dataset of images. We not only report the accuracy of the segmenter but also assess its impact on contactless fingerprint-matching accuracy.

\section{Research Methods}

In this section, we provide a detailed examination of the method used for collecting fingerphotos from adult subjects, data annotation, data augmentation, and ground truth labeling. Then, we describe the proposed neural network architecture of CRFSEG-v2. Finally, we discuss the metrics used to evaluate different slap segmentation algorithms.

\subsection{Slap Dataset}
In this section, we discuss contactless datasets including fingerphoto collection, annotation, augmentation, and ground truth creation.

\subsubsection{Contactless Fingerphoto Collection}
Contactless fingerphoto collection is a crucial component of contactless fingerprint biometrics research. Our literature review revealed a shortage of contactless fingerprint images, particularly finger photos containing multiple fingers. To address this gap, we conducted a user study and collected fingerphotos multiple times. The collection process involved acquiring fingerprint images without any physical contact with fingerprint capture devices. In this subsection, we discuss the different techniques and considerations involved in contactless fingerphoto collection.

\begin{itemize}
    \item Camera Selection: The selection of a camera is of utmost importance in contactless finger photo collection. Opting for high-resolution cameras with excellent color reproduction and minimal noise levels is ideal for capturing clear and detailed finger images. However, when developing a contactless authentication system that operates in real-world scenarios, relying solely on high-quality images is not recommended. Therefore, we utilized multiple mobile devices, including the Google Pixel 2, iPhone 7, X, Samsung S6, S7, and S9 to capture fingerphotos. 

    \item Illumination Conditions: The lighting conditions during the capture of contactless fingerphotos are crucial. It is important to have sufficient and uniform illumination to minimize shadows, reflections, and image noise. We took precautions to avoid overexposure or underexposure, as these can affect the visibility of fingerprint details.

    \item Finger Alignment and Placement: Providing clear instructions and visual cues can help users achieve optimal finger placement. However, for real-world applications, fingerprint alignment and placement can vary. Therefore, we did not instruct users on how to position their fingers specifically within the camera's field of view. As a result, we obtained fingerprints with diverse alignment and placement. 

    \item Motion Blur Reduction: Movement during image capture can cause motion blur, which can degrade the quality of finger images. To minimize motion blur and improve image sharpness, users were instructed to place their hands on a table or stable surface.

   \item Backgrounds: Selecting an appropriate background is crucial for improving the contrast between the finger and the background, which facilitates accurate extraction of the finger region. However, it is not always possible for users to have an ideal background in real-world applications. Therefore, we employed various backgrounds when capturing fingerphotos.

   \item Privacy and Data Protection: Collecting contactless fingerphotos involves capturing personal biometric data. It is crucial to comply with privacy regulations and ensure the secure storage and handling of the collected data. To protect the privacy and confidentiality of individuals' biometric information, we employed anonymization techniques, and with appropriate permits from the Institutional Review Board (IRB).
\end{itemize}

This dataset comprises a total of 2150 fingerphotos. A comprehensive and diverse image dataset should encompass a broad range of scenarios, including different poses, illumination conditions, sizes, brightness levels, and fingerprint positions. Such datasets are valuable for developing a robust deep-learning-based fingerprint segmentation model and are compiled to thoroughly evaluate the model's performance in real-world scenarios. To introduce such variation into our dataset, we employed data augmentation techniques. Through augmentation, we obtained an additional 21,500 augmented images, thereby expanding the dataset to a total 23,650 images.
\subsection{Fingerphoto Annotation and Augmentation}
Data annotation plays a crucial role in developing a well-structured and representative dataset, which is essential for building a reliable deep learning-based image segmentation model. The annotation process involves delineating a bounding box around each fingertip in a fingerphoto and assigning a corresponding label to each fingertip. The fingertip labels used include Left-Index, Left-Middle, Left-Ring, Left-Little, Left-Thumb, Right-Index, Right-Middle, Right-Ring, Right-Little, and Right-Thumb. This annotated data is then utilized to train a deep learning model, enabling it to recognize and classify distinct objects or classes within a fingerphoto.

The creation of a ground truth dataset for evaluating contactless fingerprint segmentation models typically involves laborious manual annotation, especially when dealing with a large number of images. To deal with this process, we initially annotated 200 images and fine-tuned our previously developed contact-based fingerprint segmentation model using this annotated dataset. Subsequently, we employed the fine-tuned CRFSEG-v2 segmentation model to automatically segment the fingerphotos. However, a manual review of the CRFSEG-v2 results was still necessary to ensure accuracy and alignment. This involved visually examining the images, rectifying errors, and aligning the fingerprints. Although the CRFSEG-v2 model demonstrated proficiency in generating bounding boxes around fingertips, it occasionally mislabeled them. We rectified these misclassifications through manual inspection of all the fingerphotos. This automated segmentation approach resulted in a 65-70\% reduction in the time required for annotation. We further augmented the contactless dataset by rotating all the fingerphotos at various angles (-90° to 90°) to create a diverse set of slap images containing rotated fingerprints. The details of the dataset are shown in the \autoref{tab:dataset_info}.
\begin{table}[htbp]
    \centering        
    \caption{We have 23,650 finger photos. Out of these, 2,150 were collected by Clarkson University. For image collection, we used different types of mobile phones such as Samsung S20, iPhone 7, iPhone X, and Google Pixel. Subsequently, we annotated the images manually. To create rotating slap images, we utilized the 2,150 finger photos that were annotated by humans to create more images using an augmentation technique. This resulted in the generation of 21,500 more augmented images by rotating all the finger photos at different angles (-90 to 90 degrees). This augmentation is intended to help the model become invariant to different types of rotations of finger photos.}
    \begin{tabular}{|l|c|c|c|}
        \hline
        Dataset & Total fingerphotos & Lefthand fingerphotos & Righthand fingerphotos \\
        \hline
        Bonafide & 2150 & 1118 & 1032 \\
        \hline
        Augmented & 21500 & 11180 & 10320 \\
        \hline
        Total & 23650 & 12298 & 11352 \\
        \hline
    \end{tabular}

    \label{tab:dataset_info}
\end{table}

\subsection{Deep Learning Architecture for Contactless fingerphoto Segmentation}

In prior work, we developed a two-stage Faster R-CNN architecture for segmenting contact-based slap fingerprints \cite{murshed2023deep}. We utilized a similar two-stage faster architecture for segmenting contactless slap fingerphotos. However, contact-based and contactless fingerprint images possess distinct characteristics in terms of quality, structure, size, and other factors.
Contact-based images tend to exhibit higher quality as they are captured using physical sensors that make direct contact with the finger. Moreover, these images benefit from stable lighting conditions. In contrast, contactless images often suffer from challenges such as blurriness, noise, diverse backgrounds, and variations in illumination, as they are captured from a distance using different types of sensors. Consequently, we needed to make small modifications to the architecture to accommodate the lower quality and address other difficulties inherent in contactless images. This modification involves the use of different backbone networks, loss functions, increasing the kernel size, and adjusting the learning rate and number of epochs.
The contactless fingerphoto deep learning architecture, see \autoref{fig:rotatedFasterRCNNArchi}, comprises three key structural components: the box head, the oriented region proposal network, and the backbone network.
\subsubsection{Backbone Network}
CRFSEG-v2 backbone network, featuring the Resnet-100 with FPN architecture, is designed to extract feature maps of different sizes from contactless fingerphotos. This network incorporates both stem blocks and bottleneck blocks. To optimize calculations, extract key features, and expand channel capabilities, bottleneck blocks with three convolution layers of varying kernel sizes (1$\times$1, 3$\times$3, and 1$\times$1) are employed \cite{He2016RESNET}. Within the stem block, the focus is on reducing the input image size, achieved through Convolution 2D layers, ReLU activation, and max-pooling layers, to minimize computational cost while retaining all essential information. The oriented region proposal network (O-RPN) then takes in the multiscale semantic-rich feature maps generated by the backbone network.

\subsubsection{Oriented Region Proposal Network (O-RPN)}
After being obtained from the backbone network, the input image feature maps are employed by the Oriented Region Proposal Network (O-RPN) to generate proposals for oriented regions of interest (ROIs), which are most likely to encompass the objects of interest. This is achieved by identifying areas with a high probability of containing the target objects.
The proposal of ROIs by the O-RPN is a crucial step as it helps in generating accurate rotated bounding boxes for both axis-aligned and rotated objects. This is in contrast to the conventional regional proposal network (RPN) used in the Faster R-CNN architecture, where only axis-aligned regions are proposed.

The proposal of oriented regions is achieved by using a sliding window approach with a 3$\times$3 kernel on the feature maps. Consequently, the system generates anchor boxes with diverse aspect ratios, scales, and orientations. If we have $k_a$ different orientations, $k_s$ different scales, and $k_r$ different aspect rations, then  $K = k_a \times k_s \times k_r $ anchor boxes are generated for each position in the feature map. Most of these anchor boxes might not have targeted objects in them. Subsequently, convolution layers and parallel output layers called localizing and classifying layers are employed to differentiate anchor boxes with target objects from others. The classifying layer assigns foreground or background labels according to the intersection-over-union (IoU) score with ground truth boxes while learning offsets (x,y,w,h, $\theta$) is done by the localization layer for the foreground boxes. The term "Foreground" is used to denote regions of interest (proposals) that exhibit a substantial overlap with the bounding boxes of the ground truth objects in the image. In simpler terms, these are areas most likely to contain a prominent object. The classifying layer assigns a "foreground" label to these regions. Conversely, regions of interest that lack a significant overlap with any ground truth bounding box are designated as the "background". The probability of an object being present in these areas is lower. The classifying layer assigns a "background" label to these regions.

\begin{figure*}[t]
 \includegraphics[width=1\linewidth]{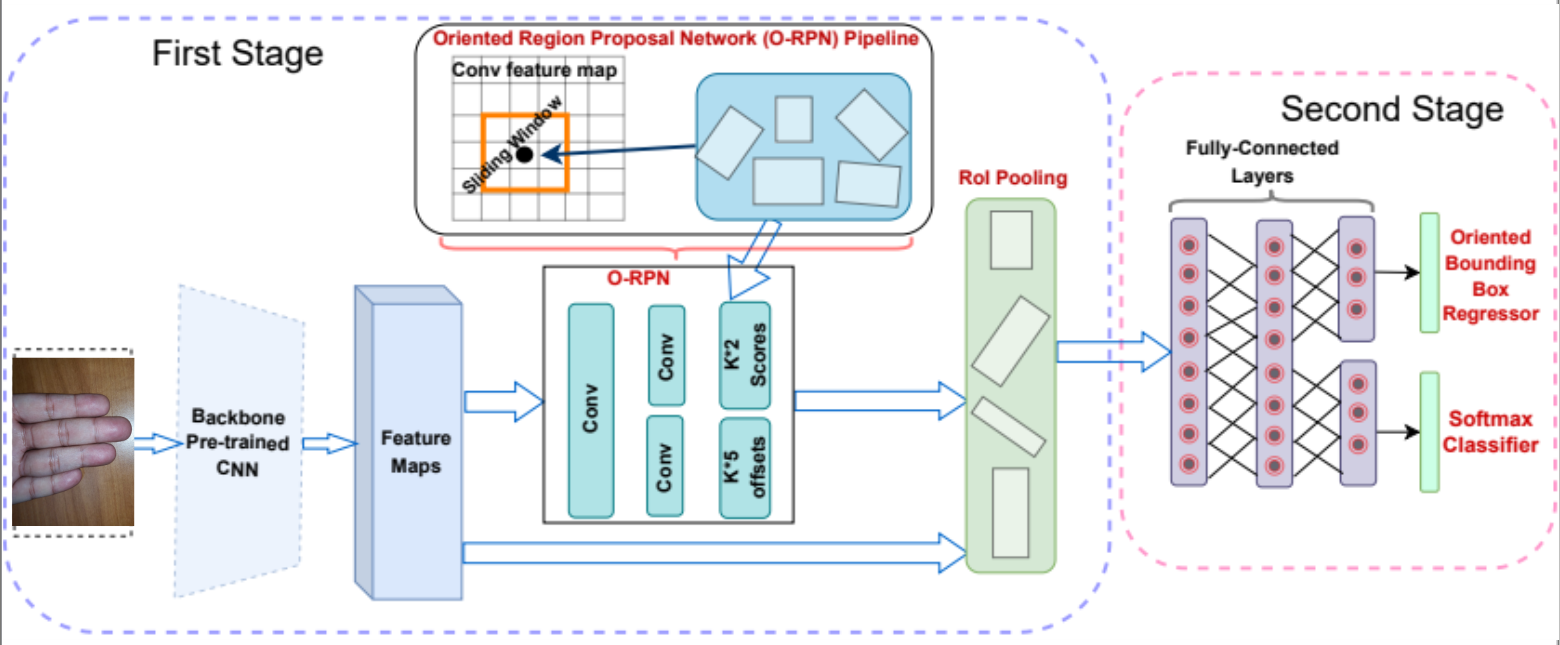}
 \caption{The complete architecture for CRFSEG-v2 includes several processing stages and is intended for precise fingerprint segmentation. For the feature maps, we need to provide our input contactless fingerprint image to a pre-trained CNN model on the ImageNet dataset. To generate oriented anchors, O-RPN needs to run on all levels of feature maps. By selecting spatial features from O-RPN and from the output of the backbone network, ROI pooling layers generate fixed-length feature vectors. These fixed-length feature vectors are then passed through the fully connected layers. There are two parallel branches that receive the output of the fully connected layers, referred to as the Softmax Classifier and Oriented Bounding Box Regressor. The softmax layers contain a softmax layer for multiclass classification, and the regressors contain bounding box regression.}
 \label{fig:rotatedFasterRCNNArchi}
\end{figure*}

Regarding the regression offset, the localizing layer generates (K$\times$5) parameterized encodings, while the classifying layer produces (K$\times$5) parameterized scores for region classification. The anchor generation strategy is illustrated in \autoref{fig:rotatedFasterRCNNArchi}.

For training O-RPN, seven different orientations (-$\pi/4$, -$\pi$/6, -$\pi$/12, 0, $\pi$/12, $\pi$/6, $\pi$/4), three aspect ratios (1:1, 1:2, 2:1 ) and three scales (128, 256 and 512) are used to generate anchors. In conclusion, the selection of seven orientations, three aspect ratios, and three scales involves a trade-off between maintaining computational efficiency during training and inference while ensuring the model's capacity to capture the diversity of objects in images \cite{jia2019embdn}. Then, all the anchors are divided into three categories 
\begin{itemize}
    \item positive anchors: the Intersection over Union (IoU) overlap between these anchors and the ground-truth boxes exceeds 0.7, indicating a strong alignment with the target objects, 
    \item negative anchors: these anchors have an IoU overlap smaller than 0.3 with the ground-truth boxes, indicating a significant mismatch with the target objects,
    \item neutral anchors: these anchors have an IoU overlap between 0.3 and 0.7 using the ground-truth boxes. These are removed from the anchor set and not used during subsequent processing.
\end{itemize}
In contrast to Faster R-CNN, where horizontal anchor boxes are used, our approach utilizes oriented anchor boxes and oriented ground-truth boxes. The loss functions employed to train the O-RPN are defined by the following equations:
\begin{equation}
    L_{o-rpn} = L_{cls}(p,u) + \lambda u L_{reg}(t, t^*) 
\end{equation}
Here, $L_{cls}$ is the classification loss, $p$ is the predicted probability across the foreground and background classes by the softmax function, $u$ represents class label for anchors, where u = 1 for foreground containing fingerprint and u = 0 for background; $t = (t_x, t_y, t_h, t_w, t_\theta)$ denotes the predicted regression offset value of an anchor calculated by the network, and $t^* = (t^*_x, t^*_y, t^*_h, t^*_w, t^*_\theta)$ represents ground truth. $\lambda$ is a balancing parameter that manages the balance between class loss and regression loss. Only the regression loss is enabled if u = 1 for the foreground and there is no regression for the background.
The classification loss function is defined as the cross-entropy loss between the ground-truth label $u$ and the predicted probability $p$:
\begin{equation}
    L_{cls}(p,u) = -u \cdot \log(p) - (1-u) \cdot \log(1-p)
\end{equation}
Tuple $t$ and $t^*$ are calculated like this:
\begin{equation}
    \begin{split}
    t_x = (x - x_a)/w_a, t_y = (y-y_a)/h_a, \\
    t_w = \log (w/w_a), t_h = \log(h/h_a),\\
    t_\theta = \theta-\theta_a
    \end{split}
\end{equation}
\begin{equation}
    \begin{split}
    t^*_x = (x^* - x_a)/w_a, t^*_y = (y^*-y_a)/h_a,\\ 
    t^*_w = \log (w^*/w_a), t^*_h = \log(h^*/h_a),\\
    t^*_\theta = \theta^*-\theta_a
    \end{split}
\end{equation}
Where $x$, $x_a$ and $x^*$ denote predicted box, anchor and
ground truth box, respectively; similar for $y$, $h$, $w$ and $\theta$.
The smooth-L1 loss is adopted for bounding box regression as follows:
\begin{equation}
    L_{reg}(t, t*) = \sum_{i \in {x,y,w,h,\theta}} u.\text{smooth}_{L1} (t^*_i - t_i)
\end{equation}
\begin{equation}
    \text{smooth}_{L1} (x) = \left\{
    \begin{array}{ll}
        {0.5x^2} & \mbox  {if |x| <1} \\
        {|x| - 0.5} & \mbox {\text{otherwise}}
    \end{array}
\right. 
\end{equation}
\subsubsection{Box Head}
The O-RPN network, as part of the architecture, generates 1000 proposal boxes and objectless logits by default. These proposal boxes are then projected into the feature space using an ROI pooling layer. The output of the ROI pooling layer is reshaped and fed into the fully connected (FC) layers. The FC layers generate a RoI vector, which is then passed through a predictor with two branches: the rotated bounding box regressor and the classifier.
The classifier, located in the classification layer of the model, predicts the object class, while the regressor layer is responsible for regressing the bounding box values. This process enables the model to accurately identify and classify objects within the input image, and also refine the location of the proposed bounding boxes.
\subsection{Evaluation Metrics}
We used Mean Absolute Error, Error in Angle Prediction, fingerprint labeling accuracy, and matching score to assess the effectiveness of the CRFSEG-v2 model on our contactless fingerphoto dataset.  

\subsubsection{Mean Absolute Error}

The Mean Absolute Error is a metric applied to evaluate the performance of fingerprint segmentation models in accurately segmenting fingerprints within a specified geometric tolerance compared to human-annotated ground truth data.

Fingerprint segmentation models need to strike a balance between two factors: over-segmentation and under-segmentation. Over-segmentation occurs when the predicted bounding box is smaller than the ground truth fingerprint area, potentially capturing the ridge-valley structure of adjacent fingerprints and introducing noise that may impact matching performance. Under-segmentation, on the other hand, involves extending the predicted bounding box beyond the actual fingerprint area, resulting in the loss of valuable fingerprint details and degrading matching performance. The MAE metric helps quantify the extent of over-segmentation or under-segmentation produced by the model. To calculate the Mean Absolute Error, we measure the distance in pixels between each side of the predicted bounding box and the corresponding side of the annotated ground truth bounding box. 
A successful segmentation refers to finding a bounding box around a fingerprint within a certain geometric tolerance of the human-annotated ground truth bounding box.

    \begin{figure*}[t]
    \centering
    \includegraphics[width=0.5\linewidth]{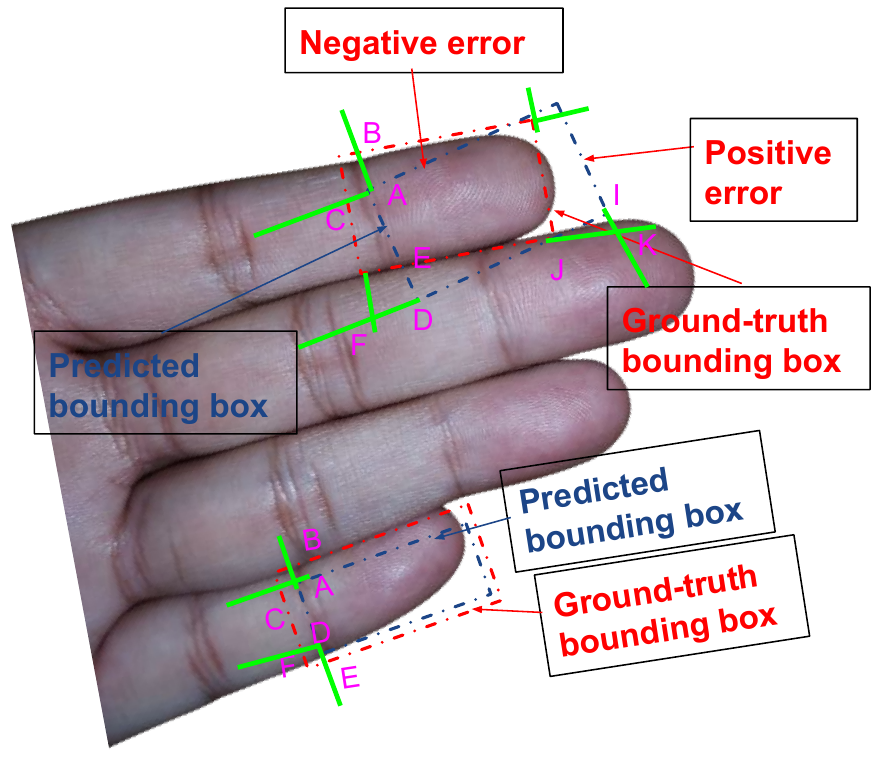}
     \caption{An example of calculating the positive and negative errors between the predicted and ground truth bounding boxes involves using Euclidean distance. To determine the pixel error for each side of the predicted bounding boxes, the distances between the endpoints of a side and the corresponding endpoints of the annotated ground-truth bounding box are computed in pixels. For instance, when calculating the pixel error for the side AD, perpendicular lines AC and DF are drawn with respect to the line AD. The perpendicular line AC intersects the corresponding ground-truth line at point C, while the perpendicular line DF intersects the extension of the corresponding ground-truth line at point F. Subsequently, the Euclidean distances from point A to point C and from point D to point F are calculated. The average of these two Euclidean distances represents the pixel error for the side AD. This process is repeated for all four sides of the bounding box. Finally, \autoref{equ:ch8_MAEs} is individually applied to the pixel errors of each side, resulting in the calculation of the Mean Absolute Error for that specific side.}
     \label{fig:Ch8_MAECalculation}
    \end{figure*}

    \autoref{fig:Ch8_MAECalculation} illustrates more detail about calculating the MAE for a fingerprint. A detected bounding box is considered to have a positive error if any of its sides encompass more data than the corresponding side of the ground-truth bounding box. An example of a positive error is shown on the right side of the rightmost fingerprint. A detected bounding box is considered to have a negative error if any of its sides enclose less data than the corresponding side of the ground-truth bounding box. An example of the negative error is shown on the top side of the left-most fingerprint.

 Finally, \autoref{equ:ch8_MAEs} is applied independently to calculate the MAE for each side.    
    \begin{equation}
    \label{equ:ch8_MAEs}
        \text{MAE} = \frac{1}{N}{\sum_{i = 0} ^ {N} |X \: error_i|}
    \end{equation}
    $N$ represents the total number of fingerprints within the dataset under evaluation. $X$ signifies the Euclidean distance error on any side of the bounding box, encompassing the top, bottom, left, and right sides.

\subsubsection{Error in Angle Prediction} Error in Angle (EAP) Prediction is employed to assess the ability of various fingerprint segmentation systems to accurately predict the orientation of fingerprints. This involves determining the deviation between the predicted angle of a fingerphoto and the ground truth angle of the same fingerphoto, which is calculated using \autoref{equ:ch8_EAP}.
    \begin{equation}
    \label{equ:ch8_EAP}
        \text{EAP} = \frac{1}N{\sum_{i = 0} ^ {N} |\theta - \theta^*|}
    \end{equation}
    Where $N$ is the total number of fingerprints in the test dataset, $\theta$ is a ground truth angle and $\theta^*$ is a predicted angle by a fingerprint segmentation model. 
    
\subsubsection{Fingerprint classification accuracy}
 Hamming loss is employed to assess the performance of multi-class classifiers \cite{Tsoumakas2007MultiLabelCA}. A pair of sequences, one comprising the ground-truth label and the other containing the predicted label, is evaluated using hamming loss, which quantifies the number of positions where the corresponding symbols differ. \autoref{equ:HLequation} is used to calculate the hamming loss \cite{Tsoumakas2007MultiLabelCA}.
    \begin{equation}
        \label{equ:HLequation}
        \text{Hamming loss} = \frac{1}{N} \sum_{i = 1} ^ {N} \frac{|Y_i \Delta Z_i|}{|L|}
    \end{equation}
    where $N$ is total number of samples in dataset, and $L$ is number of labels. $Z_i$ is the predicted value for the i-th label of a given sample, and $Y_i$ is the corresponding ground true value. $\Delta$ stands for the symmetric difference between two sets of predicted and ground truth values.

    The accuracy of a multi-class classifier is related to Hamming loss \cite{ha2021topic, koyejo2015consistent}, which can be computed using \autoref{equ:ch8_AccuracyHL}. 
    \begin{equation}
    \label{equ:ch8_AccuracyHL}
        \text{Accuracy} = 1 - \text{Hamming loss}
    \end{equation}

\subsubsection{ Fingerprint Matching}
Fingerprint matching is a crucial aspect of slap segmentation systems, where the performance of the algorithms in correctly segmenting fingerprints within a specific tolerance is evaluated. To assess fingerprint matching, we rely on the true accept rate (TAR) and false accept rate (FAR). The TAR represents the percentage of instances in which a biometric recognition system accurately verifies an authorized individual, calculated using \autoref{equ:ch8_TAR}:
    \begin{multline}
       \label{equ:ch8_TAR}    
        TAR = \frac{\text{Correct accepted fingerprints}}{\text{Total number of mated matching attempts}} \times 100\%
    \end{multline}

The false accept rate (FAR) measures the percentage of instances in which a biometric recognition system mistakenly verifies an unauthorized user. It is computed using \autoref{equ:ch8_FAR}.
    
    \begin{multline}
    \label{equ:ch8_FAR}
        FAR = \frac{\text{Wrongly accepted fingerprints}}{\text{Total number of non-mated matching attempts}}
        \times 100\%
    \end{multline}

\subsection{Training}
The CRFSEG-v2 model was developed using Detectron2, a framework created by Facebook AI Research (FAIR) \cite{wu2019detectron2}, which supports advanced deep learning-based object detection algorithms. We utilized the Faster R-CNN algorithm and customized the Detectron2 code to implement the oriented regional proposal network (ORPN), added new layers for handling rotated bounding boxes, to meet our specific requirements for accurate slap fingerprint segmentation.\par 
In our experiments, we started with a pre-trained Faster R-CNN model trained on the MS-COCO dataset, which has 81 output classes. However, since our task involved classifying ten fingerprints from two hands, we adjusted the output layers to reduce the number of classes from 81 to 10. The model was then fine-tuned using our unique slap image dataset.
Training followed an end-to-end strategy, where we calculated loss values by comparing the predicted results against the ground truth. The training was conducted on a Linux-operated machine equipped with a 20-core Intel(R) Xeon(R) E5-2690 v2 @ 3.00GHz CPU, 64 GB RAM, and a NVIDIA GeForce 1080 Ti 12-GB GPU.

We performed a total of 40,000 training iterations for the fingerprint segmentation model. The learning rates started at $10^{-4}$ and were decreased by a ratio of 0.1 at specific intervals (4000, 8000, 12000, 18000, and 25000, 32000 iterations). The weight decay was set to 0.0005, and the momentum was set to 0.7. Throughout the experiments, we employed multi-scale training, eliminating the need for scaling the input before feeding it into the neural network.
The contactless fingerprint dataset, consisting of 23,650 finger photos, is divided using an 80:10:10 train/validate/test split ratio. A 10-fold cross-validation technique is employed to construct and assess the model, and the outcomes are presented in the results section.

\section{Results}
This section presents a comprehensive analysis of our findings. We employed four distinct metrics, namely Mean Absolute Error, Error in Angle Prediction, fingerprint labeling accuracy, and fingerprint matching accuracy, to assess the performance of our nobel CRFSEG-v2 segmentation model that handles contacless (contactbased) fingerprint images.

\subsection{Mean Absolute Error of CRFSEG-v2 on our Contactless dataset}
MAE measures the preciseness of bounding boxes around fingerprints generated by slap segmentation algorithms.
We used \autoref{equ:ch8_MAEs} to calculate the MAE of our contactless fingerphoto segmentation model on our novel dataset.

\autoref{table:ch8_AvgMAE} presents the Mean Absolute Error and its corresponding standard deviation for the segmentation model. The MAEs obtained for different sides of the predicted bounding boxes are as follows: 26.09 for the left side, 27.33 for the right side, 20.23 for the top side, and 52.92 for the bottom side. It is worth noting that all these values are below the NIST-defined tolerance threshold of 64 pixels. Furthermore, the achieved MAEs are comparable to those obtained by the contact-based fingerprint segmentation system, indicating the effectiveness of the proposed approach in achieving accurate segmentation results.

\begin{table}[htbp]
\centering
\caption{The Mean Absolute Error (MAE) and its standard deviation were computed to assess the performance of the contactless segmentation system on our contactless fingerphoto dataset. The MAE was determined by averaging the absolute differences between each side of the detected bounding box and the corresponding side of the ground-truth bounding box, measured in pixels. A lower MAE value indicates better performance in accurately segmenting the fingerphotos.}
\begin{tabular}{|c|c|c|}
\hline
Dataset & Side & MAE(Std. dev.) \\
\hline
\multirow {4}{*}{Contactless} & Left & 26.09 (65.36) \\
\cline{2-3}
{} & Right & 27.33 (64.29) \\
\cline{2-3}
{} & Top & 20.23 (52.97) \\
\cline{2-3}
{} & Bottom & 52.92 (90.93) \\
\hline
\end{tabular}
\label{table:ch8_AvgMAE}
\end{table}

\begin{figure}[htbp]
    \begin{subfigure}[b]{0.5\textwidth}
    \centering
    \includegraphics[width=1\linewidth]{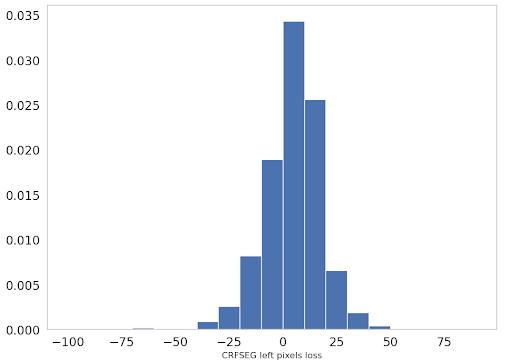}
    \caption{The Mean Absolute Error (MAE) in pixels for the left side of the fingerprints, as predicted by the contactless segmentation system.}
    \label{fig:Ch8_MAE_Left}
    \end{subfigure}
    \begin{subfigure}[b]{0.5\textwidth}
    \centering
    \includegraphics[width=1\linewidth]{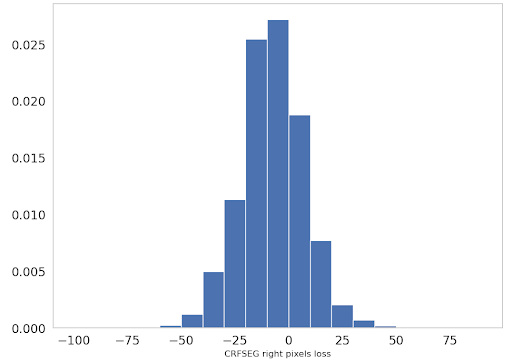}
    \caption{The Mean Absolute Error (MAE) in pixels for the right side of the fingerprints, as predicted by the contactless segmentation system.}
    \label{fig:Ch8_MAE_Right}
    \end{subfigure}
    \begin{subfigure}[b]{0.5\textwidth}
    \centering
    \includegraphics[width=1\linewidth]{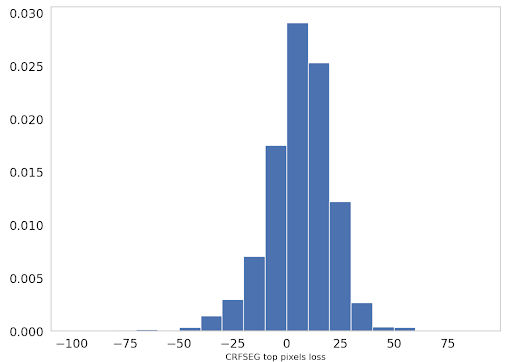}
    \caption{The Mean Absolute Error (MAE) in pixels for the top side of the fingerprints, as predicted by the contactless segmentation system.}
    \label{fig:Ch8_MAE_Top}
    \end{subfigure}
    \begin{subfigure}[b]{0.5\textwidth}
    \centering
    \includegraphics[width=1\linewidth]{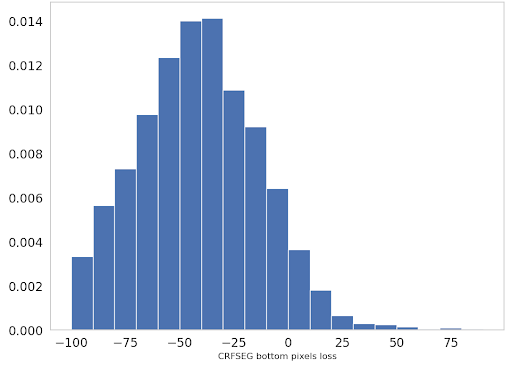}
    \caption{The Mean Absolute Error (MAE) in pixels for the bottom side of the fingerprints, as predicted by the contactless segmentation system.}
    \label{fig:Ch8_MAE_Bottom}
    \end{subfigure}
    \caption{The MAE histograms obtained from contactless fingerphoto segmentation algorithm using \autoref{equ:ch8_MAEs} are presented in this figure. \autoref{fig:Ch8_MAE_Left}, \autoref{fig:Ch8_MAE_Right}, \autoref{fig:Ch8_MAE_Top}, and \autoref{fig:Ch8_MAE_Bottom} illustrate the MAE values for the left, right, top, and bottom sides of the bounding box, respectively, as predicted by the contactless fingerphoto segmentation system.}
  \label{fig:ch8_MAEFullHits}
\end{figure}

\autoref{fig:ch8_MAEFullHits} showcases the histograms illustrating the Mean Absolute Error for all sides of the bounding boxes generated by our contactless segmentation model. These histograms are employed to showcase, analyze, and evaluate the MAE results. The generation process involved subtracting the coordinate positions of the corresponding sides of the ground-truth bounding boxes from the coordinate positions of the corresponding sides of the detected bounding boxes obtained from the segmentation model. The histograms provide evidence of improved performance in accurately segmenting contactless fingerphotos.

\subsubsection{Error in Angle Prediction}
The Error in Angle Prediction (EAP) refers to the discrepancy between the angles predicted by the segmentation algorithms and the corresponding ground-truth angles of fingerphotos. It directly reflects the precision of the bounding boxes detected by the segmentation model. A smaller deviation among the predicted angles and the ground-truth angles signifies more accurate segmentation outcomes. The calculation of EAP is performed using \autoref{equ:ch8_EAP}.

For our contactless segmentation system, the resulting EAP is $5.92^\circ$, accompanied by a standard deviation of $11.98^\circ$. The EAP values obtained from the segmentation model are comparably lower and exhibit a similar trend to those of the contact-based slap fingerprint segmentation, indicating superior performance. 

\begin{figure*}[htbp]
  \centering
  \includegraphics[width=0.5\linewidth]{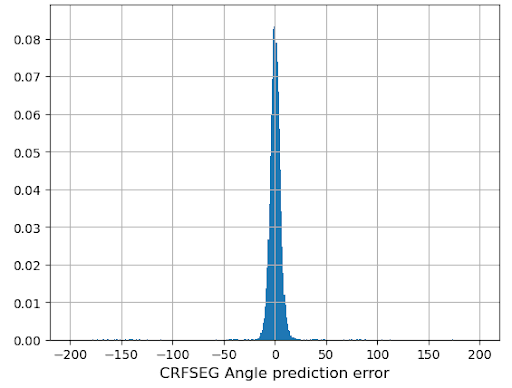}
  \caption{The histogram of the error in fingerphoto angle prediction by the contactless segmentation model on the fingerphoto dataset. This error is calculated by subtracting the angles predicted by the models from the ground-truth angles of the fingerphoto images. Low standard deviation values indicate better results.}
  \label{fig:ch8_AvgAngleErrorhis}
\end{figure*}

\autoref{fig:ch8_AvgAngleErrorhis} displays the histogram of the EAP generated by the contactless segmentation model. The narrower spread of the histogram, as indicated by the smaller standard deviation and area, demonstrates the superior performance of the segmentation model in angle prediction.

\begin{figure*}[t]
\centering
\includegraphics[width=0.7\linewidth]{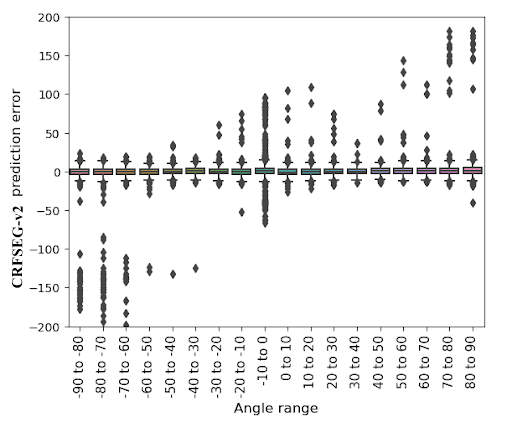}
\caption{Boxplots of the error in angle prediction (EAP) of the contactless segmentation model on the contactless fingerphoto dataset. The boxplot statistical analysis of the mean with
 $\pm 10^\circ$ (standard errors) for the EAP values of the model indicates that the algorithm used in this model is invariant to the rotation of fingerphotos.}
 \label{fig:ch8_boxplot_angle_error}
\end{figure*}

The boxplot graph in Figure \ref{fig:ch8_boxplot_angle_error} displays the EAP values obtained from the contactless segmentation model, illustrating the minimum, lower quartile, median, upper quartile, and maximum values. The line extending across the box corresponds to the median value of the EAP.
Upon examining the boxplot graph, it becomes evident that the absolute median value of the EAP generated by the contactless segmentation model is notably lower. This indicates that the model demonstrates enhanced accuracy and precision in predicting the angle of fingerphotos, even when they are excessively rotated. The boxplot graph illustrates the constrained variability of the EAP values, further affirming the model's ability to maintain consistent performance regardless of the degree of rotation.  

\subsubsection{Label Prediction Accuracy of the contactless segmentation model}
The fingerprint label prediction accuracy of the contactless segmentation model was calculated using the Hamming Loss metric, which is widely used for evaluating multi-label classifiers \cite{Khamis15WTmultilebel}. Our segmentation model demonstrated a remarkable label accuracy of $97.46\%$ on our contactless dataset, underscoring its superior accuracy in predicting fingerprint labels. The model's ability to accurately predict labels solely based on the characteristics of the fingerprint images adds to its versatility and potential for broader application.

\subsubsection{Fingerprint Matching}

The main objective of a fingerprint segmentation algorithm is to improve the matching performance. In our research, we specifically focus on accurately segmenting fingerphotos to achieve higher matching accuracy. After segmenting contactless fingerprints, we applied various enhancement techniques. Utilizing the Verifinger 12.0 SDK, minutiae representations are simultaneously retrieved from the contactless images \cite{grosz2021c2cl}. We applied operations, such as the Contrast Limited Adaptive Histogram Equalization (CLAHE) technique, to enhance our contactless segmented fingerprint images \cite{chakraverti2023noising}. To measure the matching accuracy, we used the VeriFinger fingerprint matcher (version 10) which is compliant with the NIST MINEX standards \cite{watson2014fingerprint}. This allowed us to quantitatively evaluate the contributions of the segmentation algorithm to the overall matching performance.

To calculate the matching accuracy, we generated two sets of segmented fingerprint images from the contactless fingerphoto dataset. One set of segmented fingerprints is generated using the information of human-annotated (ground truth) bounding boxes and another set is generated using the information of CRFSEG-v2 model (segmentation model) generated bounding box information. We evaluated all possible genuine comparisons for each fingerprint in the segmented fingerprint set, while randomly selecting 100 non-mated fingerprints to create an imposter distribution.

In \autoref{table:ch8_MatchingAccuracy}, we present the matching performance for segmented fingerprints using both the ground-truth bounding box information and the segmentation model. Our segmentation model achieved a fingerprint-matching accuracy of 88.88\%, whereas the ground-truth accuracy was 90.36\%. However, It is crucial to remember that the fingerprint matching accuracy for both the ground-truth data and the model's output data falls short compared to the contact-based fingerprint model, which achieves an accuracy of about 99\%.
Several factors contribute to the lower matching results, which we thoroughly discuss in the discussion section. Additionally, we provide suggestions for future research to overcome these challenges and achieve better matching performance. In total, we conducted 6,208,760 comparisons across all fingers in our experiment to estimate the matching accuracy. 

\begin{table}[htbp]
\renewcommand{\arraystretch}{1}

  \centering
  \caption{The True Positive Rate (TPR) at a False Positive Rate (FPR) of 0.001 is evaluated for both ground-truth and contactless segmentation model segmented fingerprints within the dataset. The results indicate CRFSEG-v2 performed close to the ground-truth level in terms of fingerprint matching.} 
  \begin{tabular}{|c|c|}
    \hline 
   {Model} & {Accuracy} \\
   \hline
   {Ground-truth} & {90.36\%} \\
   \hline
   {Contactless Model} & {88.88\%} \\
   \hline 
\end{tabular}
  
\label{table:ch8_MatchingAccuracy} 
\end{table}

\begin{figure*}[htbp]
  \centering
  \includegraphics[width=1\linewidth]{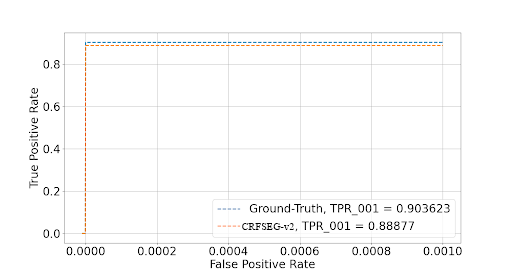}
\caption{
The Receiver Operating Characteristics (ROC) for the fingerprint matching performance of Ground Truth, a newly developed fingerphoto segmentation model in the contactless dataset.}
\label{fig:ch8_contactless_matching_results}
\end{figure*}

\autoref{fig:ch8_contactless_matching_results} is the Receiver Operating Characteristics (ROC) curve that shows the matching scores of our segmentation model CRFSEG-v2 along with ground truth on the contactless dataset. 
The ROC curve is a graphical plot that represents the tradeoff
between the true positive rate (TPR) and the false positive rate
(FPR) at various threshold values. 

\section{Discussion}
This study focuses on the development of a highly accurate slap segmentation system specifically designed for contactless fingerphoto images. The system utilizes deep learning techniques, particularly convolutional networks trained using an end-to-end approach to address challenges such as rotation and noise commonly encountered in fingerphotos. One advantage of our system is its ability to be fine-tuned using additional datasets, which enhances its performance and generalization across a wide, diverse range of fingerphoto images. To the best of our knowledge, commercial slap segmentation systems lack the flexibility to be easily fine-tuned on diverse fingerphoto datasets.

The mean absolute errors (MAEs), error in angle prediction (EAP), and labeling accuracy achieved by our segmentation model are high and comparable to the accuracy levels achieved by high-precision segmentation systems developed and tested using contact-based slap images. However, we observed lower accuracy in the matching performance of our system. It is crucial to remember that the matching performance heavily relies on the precise segmentation of fingerphotos, accurate labeling of fingertips/fingerprints, and the quality of the fingerphotos.
Our segmentation model demonstrates the ability to effectively segment fingerphotos even in cases where the image quality is poor, thereby achieving label accuracy that is nearly on par with human-level performance. However, when these segmented images were subjected to different commercial matching software, we observed a decrease in fingerprint-matching accuracy. Through manual examination, we determined that our segmentation system accurately segments fingertips and assigns accurate labels to them. We believe the reduced fingerprint matching accuracy can be attributed to the limitations of the fingerprint matching software.

\section{Conclusion}

In this paper, we have examined the potential of contactless fingerprint authentication systems as a promising alternative to traditional contact-based methods. By leveraging deep learning techniques, we have developed and evaluated a novel segmentation model for precise localization and extraction of contactless fingerprints. The results obtained from our real-world dataset demonstrate the effectiveness and reliability of our novel segmentation and extraction method.
This work centers around the development of a segmentation model that leverages deep learning techniques, specifically tailored for the segmentation of contactless fingerprints.\par  
Our novel CRFSEG-v2 model has achieved notable results, including an average Mean Absolute Error (MAE) of 30 pixels, an Error in Angle Prediction (EAP) of 5.92 degrees, a Labeling Accuracy of 97.46\%, and a VeriFinger matching accuracy of 88.87\%. Additionally, we have curated an extensive in-house contactless dataset, comprising 23,650 finger photos.
While our research showcases promising results, there are still challenges to address in the development of contactless fingerprint authentication systems. Future work should focus on addressing issues like variability in image quality, occlusion, and lighting conditions to further improve the robustness and generalization of the proposed system.
With continued research and advancements in deep learning techniques, we can expect even more sophisticated and reliable contactless fingerprint authentication systems to emerge. These developments will undoubtedly contribute to enhanced security and an improved user experience across a broad range of applications, making fingerprint authentication an indispensable part of our digital lives.

Here are some potential future steps that researchers can consider to address the matching failure and hence improve performance:

\begin{itemize}
\item GAN-based approaches for enhancing low-quality images: Developing a generative adversarial network (GAN) can help learn the mapping between low-quality fingerprint images and their corresponding high-quality representations. By training the GAN on a dataset containing pairs of low-quality and high-quality fingerprint images, it can learn to enhance the low-quality images, thereby improving the matching accuracy.

\item Feature extraction and matching algorithms: Robust feature extraction algorithms can be employed to capture relevant fingerprint information, even in the presence of variations in image quality. These algorithms should be designed to handle distortions caused by factors such as blurriness, missing parts, or poor lighting conditions. Similarly, matching algorithms should be capable of accurately comparing and aligning fingerprint features, even when dealing with imperfect images.

\item Multiple capture and fusion: Instead of relying on a single fingerprint image, multiple captures of the fingerprint can be obtained and then fused together. This approach helps mitigate issues like missing parts or blurriness in individual images. Fusion techniques such as averaging, weighted averaging, or feature-level fusion can be employed to combine the information from multiple images and improve the overall matching accuracy.
\end{itemize}

\bibliographystyle{plain} 
\bibliography{bibfile}
\section{Acknowledgements}
This material is based upon work supported by the Center for Identification Technology Research and the National Science Foundation under Grant Number 1650503.
\end{document}